\documentclass[sigconf]{acmart}
\AtBeginDocument{%
  }
\usepackage{tikz}      
\usepackage{subcaption}
\usepackage{svg}
\usepackage{xcolor}
\definecolor{thirdcolor}{RGB}{0,128,0}
\usepackage{multirow}
\usepackage{adjustbox} 
\usepackage{afterpage}
\usepackage{balance}
\usepackage{graphicx}
\usepackage{capt-of}

\setcopyright{acmlicensed}

\copyrightyear{2026}
\acmYear{2026}
\setcopyright{cc}
\setcctype{by}
\acmConference[MM '26]{Proceedings of the 34th ACM International Conference on Multimedia}{November 10--14, 2026}{Rio de Janeiro, Brazil}
\acmBooktitle{Proceedings of the 34th ACM International Conference on Multimedia (MM '26), November 10--14, 2026, Rio de Janeiro, Brazil}
\acmDOI{10.1145/3767308.3838680}
\acmISBN{979-8-4007-2213-4/2026/11}

\begin{document}

\title{VTONQA: A Multi-Dimensional Quality Assessment Dataset for Virtual Try-on}

\author{Xinyi Wei}

\email{moj-will@sjtu.edu.cn}
\authornote{These authors contributed equally to this work.}
\affiliation{
    \institution{Shanghai Jiao Tong University}
    \city{Shanghai}
    \country{China}
}

\author{Sijing Wu}

\email{wusijing@sjtu.edu.cn}
\authornotemark[1]
\authornote{Corresponding author.}
\affiliation{
    \institution{Shanghai Jiao Tong University}
    \city{Shanghai}
    \country{China}
}

\author{Zitong Xu}
\email{xuzitong@sjtu.edu.cn}
\affiliation{
    \institution{Shanghai Jiao Tong University}
    \city{Shanghai}
    \country{China}
}

\author{Yunhao Li}
\email{lyhsjtu@sjtu.edu.cn}
\affiliation{
    \institution{Shanghai Jiao Tong University}
    \city{Shanghai}
    \country{China}
}

\author{Huiyu Duan}
\email{huiyuduan@sjtu.edu.cn}
\affiliation{
    \institution{Shanghai Jiao Tong University}
    \city{Shanghai}
    \country{China}
}

\author{Jia Wang}
\email{jiawang@sjtu.edu.cn}
\affiliation{
    \institution{Shanghai Jiao Tong University}
    \city{Shanghai}
    \country{China}
}

\author{Ning Liu}
\email{ningliu@sjtu.edu.cn}
\affiliation{
    \institution{Shanghai Jiao Tong University}
    \city{Shanghai}
    \country{China}
}

\begin{abstract}
With the rapid development of e-commerce and digital fashion, image-based virtual try-on (VTON) has attracted increasing attention. However, existing VTON models often suffer from artifacts such as garment distortion and body inconsistency, highlighting the need for reliable quality evaluation of VTON-generated images. To this end, we construct \textbf{VTONQA}, the first multi-dimensional quality assessment dataset specifically designed for VTON, which contains 8,132 images generated by 11 representative VTON models, along with 24,396 mean opinion scores (MOSs) across three evaluation dimensions (\textit{i.e.}, clothing fit, body compatibility, and overall quality). Based on VTONQA, we benchmark both VTON models and a diverse set of image quality assessment (IQA) metrics, revealing the limitations of existing methods and highlighting the value of the proposed dataset. We believe that the VTONQA dataset and corresponding benchmarks will provide a solid foundation for perceptually aligned evaluation, benefiting both the development of quality assessment methods and the advancement of VTON models. The dataset we proposed in this paper is publicly available at: https://huggingface.co/datasets/weixiny0408/vtonqa.
\end{abstract}



\maketitle

\vspace{-12pt}

\begin{figure}[!t]
    \centering
    \includegraphics[width=\linewidth]{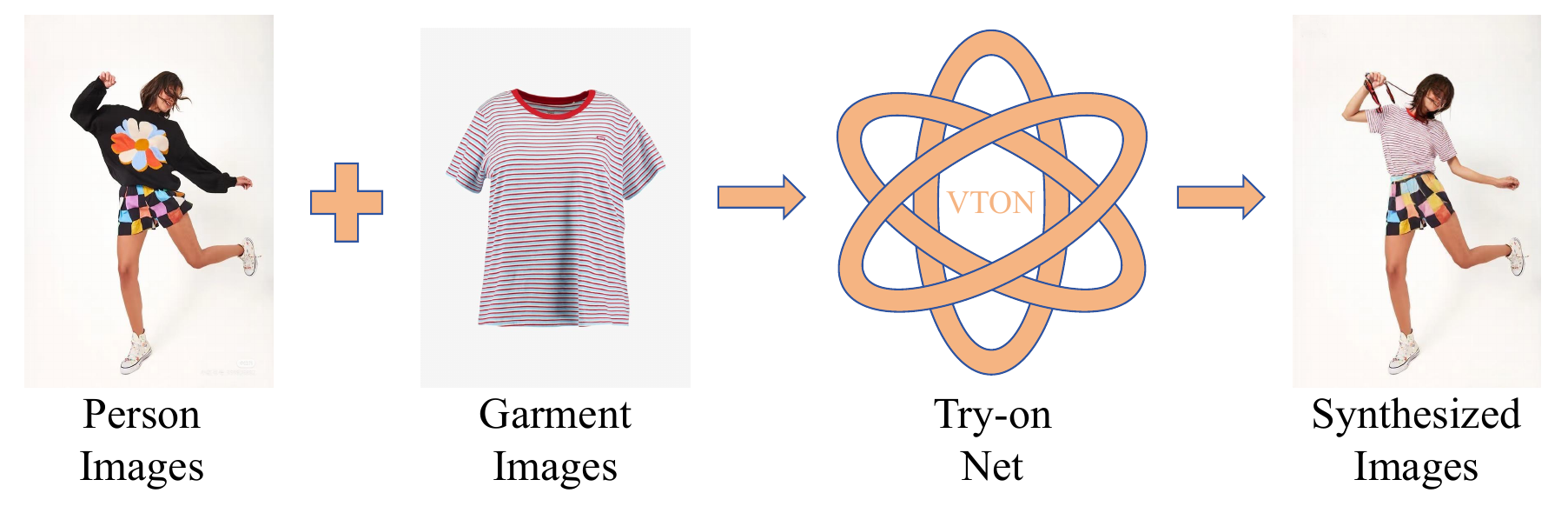}
    \vspace{-15pt}
    \caption{Illustration of the image-based virtual try-on pipeline.}
    \vspace{-12pt}
    \label{fig:pipeline}
\end{figure}

\begin{figure*}[t]
    \centering
    
    \begin{minipage}{0.48\textwidth}
        \centering
        \includegraphics[width=\linewidth]{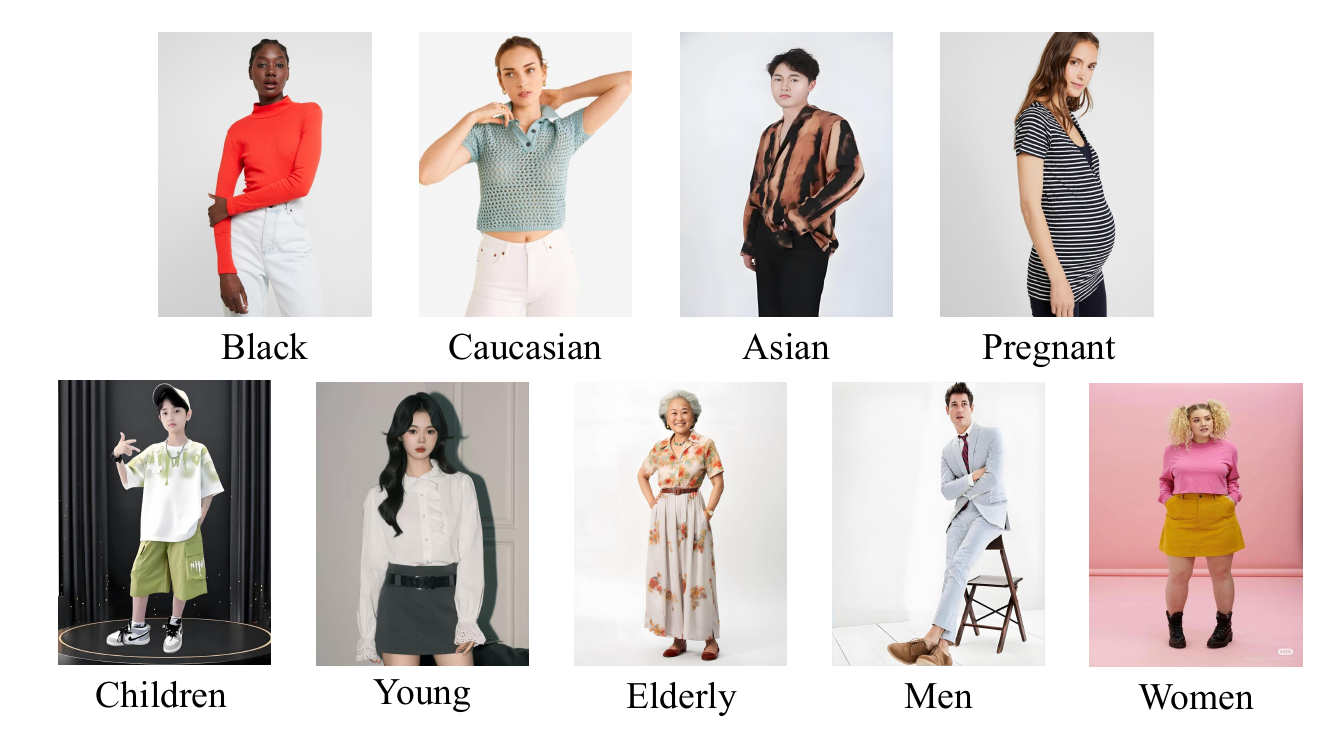}
        \vspace{-12pt}
        \caption{Representative examples of the nine reference person categories in VTONQA, illustrating the diversity of body types and dressing scenarios included in the dataset.}
        \label{fig:person_type}
    \end{minipage}
    \hfill
    \begin{minipage}{0.48\textwidth}
        \centering
        \includegraphics[width=\linewidth]{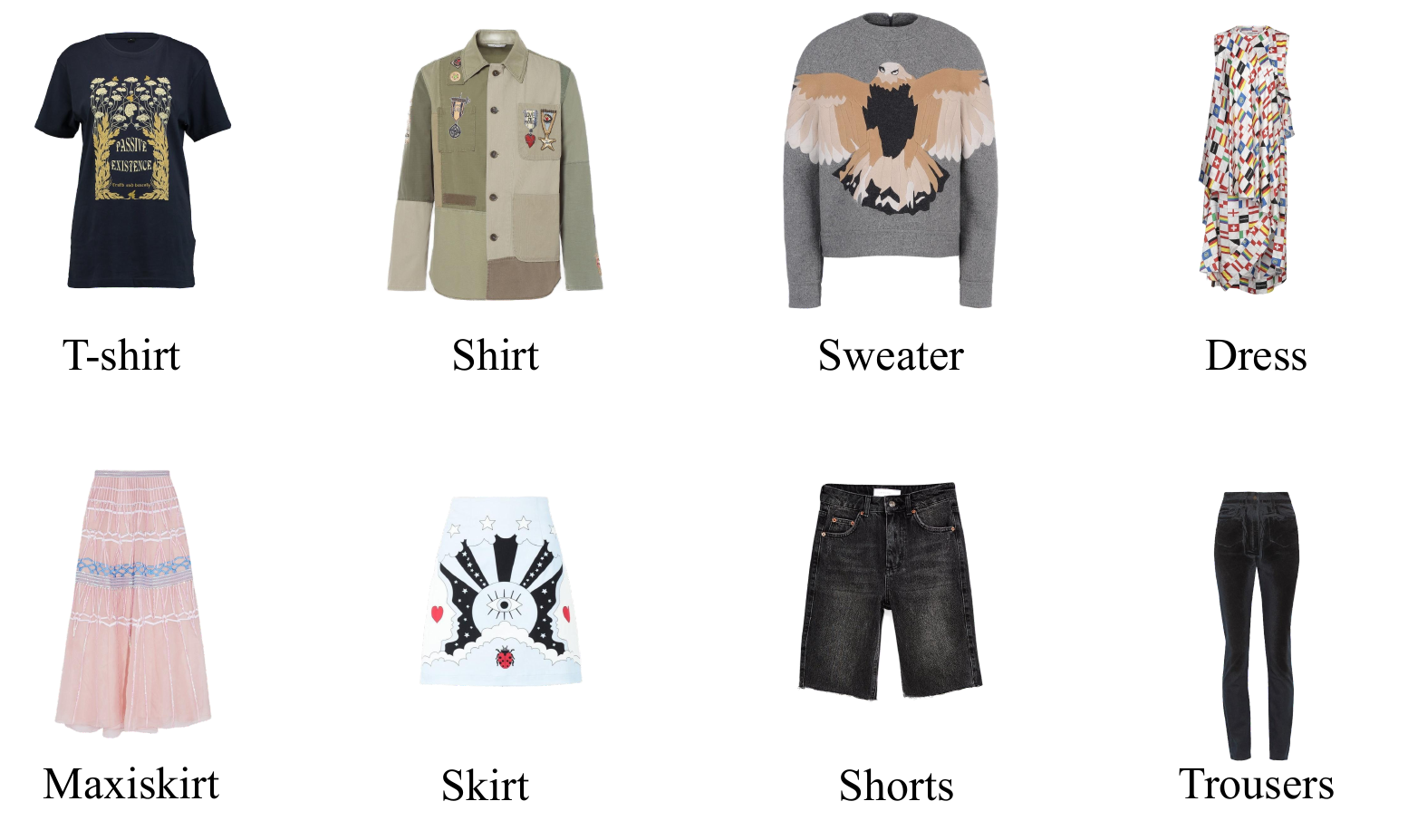}
        \vspace{-12pt}
        \caption{Representative examples of the eight clothing categories included in VTONQA, covering upper-body, lower-body, and full-body garments.}
        \label{fig:cloth_type}
    \end{minipage}
\vspace{-10pt}
\end{figure*}

\section{Introduction}

Image-based virtual try-on (VTON) \cite{choi2021viton,kim2024stableviton,Yang_2024_CVPR} enables realistic visualization of garments on human images (see Figure~\ref{fig:pipeline}), with applications in e-commerce, virtual reality, and digital fashion. However, current VTON models often generate low-quality results with artifacts such as blurred faces and garments, distorted body structures, and incorrect clothing transfer, degrading user experience and practical utility. Therefore, effective evaluation of VTON-generated images is essential for assessing perceptual quality, benchmarking VTON models, and guiding model improvement.

Existing evaluations of VTON-generated images mainly rely on objective metrics, including the distribution-based measure Fréchet Inception Distance (FID) \cite{heusel2017gans}, the perceptual similarity metric Learned Perceptual Image Patch Similarity (LPIPS) \cite{lpips}, and traditional pixel-level criteria such as Structural Similarity Index (SSIM) \cite{SSIM} and Peak Signal-to-Noise Ratio (PSNR).
However, these objective metrics often exhibit weak correlation with human perception, highlighting the importance of subjective quality assessment.
In contrast, visual quality assessment (QA) methods \cite{MANIQA,CLIPIQA,wu2025fvq,gao2025multi,li2025aghi,yang2025lmme3dhf,xu2025lmm4edit,zhou2026mi3s,duan2025bmpcqa} typically learn a network to regress quality scores based on human-annotated datasets and are inherently aligned with human perception.
However, existing QA datasets are primarily designed for natural images or specific Artificial Intelligence (AI)-generated images, and none of them specifically target the VTON task. Due to the substantial differences between VTON-generated images and natural images in distortion types, as well as differences in reference information formats and distortion characteristics compared to other AI-generated content \cite{wu2025hveval,xu2025lmm4edit,yang2025lmme3dhf}, QA models trained on these datasets often perform poorly on VTON-generated images, highlighting the need for a VTON-specific QA dataset. However, existing quality assessment methods typically provide a single holistic quality score, which is insufficient to characterize the diverse failure modes of VTON-generated images. In contrast, VTON quality depends on multiple complementary aspects, including garment transfer, body compatibility, and overall perceptual realism. Therefore, a multi-dimensional quality assessment protocol is essential for more fine-grained and interpretable evaluation.

To bridge this gap, we construct \textbf{VTONQA}, the first large-scale multi-dimensional QA dataset for VTON-generated images, comprising 8,132 images from 11 representative VTON models and 24,396 mean opinion scores (MOSs) across three dimensions: clothing fit, body compatibility, and overall quality. Specifically, the VTON-generated images are synthesized by applying try-on garments from 8 categories to 184 reference person images spanning 9 categories. 

The VTON models include classical warp-based \cite{choi2021viton,Yang_2024_CVPR,sun2025dsvtonhighqualityvirtualtryon,he2022fs_vton}, diffusion-based \cite{morelli2023ladi,zeng2024cat,xu2024ootdiffusion,kim2024stableviton,chong2025catv2tontamingdiffusiontransformers}, and closed-source methods \cite{kling2024tryon,linkfox2024aidressing}. Subsequently, 40 subjects annotate images across three evaluation dimensions, with quality ensured through standardized training and supervision by image processing researchers. Based on VTONQA, we benchmark 11 representative VTON models and 17 IQA metrics. All VTON models are evaluated in an inference-only setting without fine-tuning or retraining on VTONQA. For quality assessment, we include full-reference and no-reference IQA methods, covering traditional and deep learning approaches, and show that models trained on VTONQA achieve stronger correlations with human judgments.

We hope that VTONQA, together with its comprehensive benchmarks and subjective annotations, will facilitate the development of objective quality assessment methods for VTON-generated images that better align with human perception and advance robust, realistic VTON models.

In summary, the main contributions are: (1) We present the first comprehensive subjective quality assessment study of VTON-generated images, providing insights into human perceptual preferences across diverse try-on scenarios. (2) We build the first large-scale, multi-dimensional quality assessment dataset for VTON-generated images, containing 8,132 images and 24,396 MOS annotations across three dimensions (\textit{i.e.}, clothing fit, body compatibility, and overall quality), enabling fine-grained evaluation beyond a single score. (3) We benchmark 11 representative VTON models and 17 IQA metrics on VTONQA, revealing the limitations of existing metrics and the importance of subjective supervision.
\begin{figure*}[!t]
    \centering
    \includegraphics[width=0.9\linewidth]{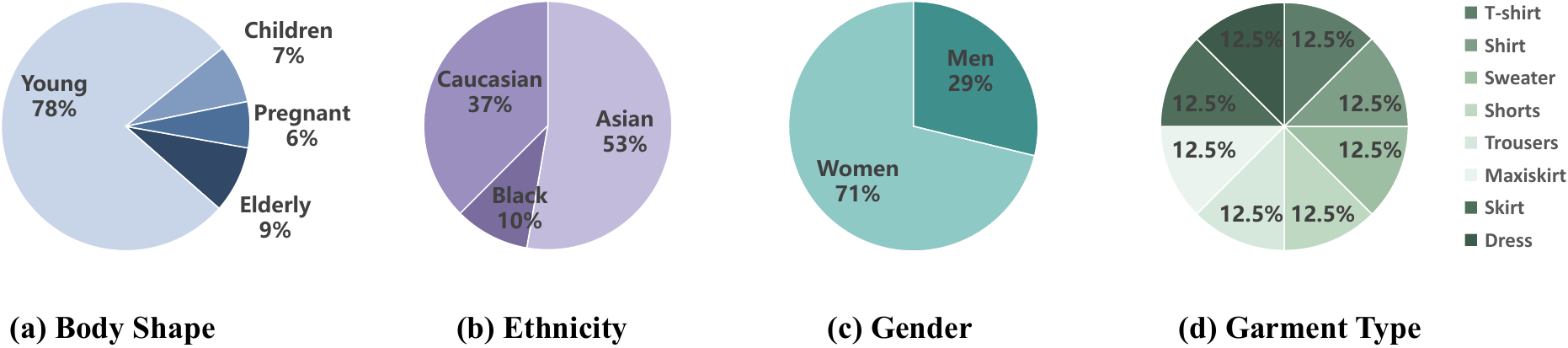}
    \vspace{-3pt}
    \caption{Distribution of garment type and demographic categories in the dataset.}
    \vspace{-8pt}
    \label{fig:distribution}
\end{figure*}

\vspace{-20pt}
\section{Related Work}
\subsection{Virtual Try-On Methods and Datasets}

Recent years have witnessed significant rapid progress in virtual try-on (VTON) research. Representative methods, such as EfficientVITON \cite{atef2025efficientviton}, CatV2TON \cite{chong2025catv2tontamingdiffusiontransformers}, and StableVITON \cite{kim2024stableviton}, improve garment detail preservation, generation quality, and inference efficiency. While early studies focused on single-view image-based try-on, recent works \cite{wang2024mv} have explored multi-view settings and 3D modeling to better capture body--garment interactions. Nevertheless, single-view VTON \cite{choi2021viton,Yang_2024_CVPR,sun2025dsvtonhighqualityvirtualtryon,he2022fs_vton,morelli2023ladi,zeng2024cat,xu2024ootdiffusion,kim2024stableviton,chong2025catv2tontamingdiffusiontransformers,kling2024tryon,linkfox2024aidressing} remains the dominant paradigm and foundation for many recent approaches. Existing VTON datasets are mainly based on VITON-HD \cite{choi2021viton} and DressCode \cite{morelli2022dresscode}. VITON-HD provides high-resolution front-view images of upper-body female subjects, while DressCode extends to full-body images, both genders, and diverse garment categories. However, both datasets still lack diversity in poses, body shapes, garment structures, and backgrounds, limiting their ability to represent real-world scenarios and comprehensively evaluate modern VTON models.

\subsection{Evaluation of Virtual Try-On Results}

Evaluating virtual try-on quality remains challenging. Existing studies mainly rely on objective image similarity metrics, including SSIM \cite{SSIM}, LPIPS \cite{lpips}, FID \cite{heusel2017gans}, and KID \cite{binkowski2018demystifying}, which measure pixel fidelity, perceptual similarity, and distributional realism. However, these metrics often fail to align with human perception in VTON scenarios involving subjective factors such as garment alignment, visual plausibility, and body–cloth interaction.

Recent works have constructed quality assessment datasets across various modalities, including image quality assessment \cite{li2025exploring,wu2026q,li2026eliq,jiang2026surveillance}, video quality and aesthetics evaluation \cite{jin2025rgc,li2026videoaesbench}, and task-specific generation scenarios such as digital humans and gesture generation \cite{li2025dhqa,gao2025ges,gao2026sfqa,wu2026multi,wu2025singinghead}. However, these datasets target generic or other generative tasks and overlook VTON-specific factors. Although some VTON works \cite{choi2021viton} provide qualitative comparisons, systematic subjective studies with quantitative ratings remain scarce, highlighting the need for perceptually aligned evaluation methods.

\section{Dataset and Evaluation Setup}

\subsection{Dataset construction}

For constructing paired data suitable for virtual try-on algorithms, we organize garments into 8 categories with 80 images: \textbf{Upper-body}: T-shirt, shirt, sweater; \textbf{Lower-body}: shorts, trousers, maxiskirt, skirt; and \textbf{Full-body}: dress. The clothing distribution is uniform, as shown in Figure~\ref{fig:distribution}(d).

Human subjects are divided into 9 demographic categories, including \textbf{Black, Caucasian, Asian, children, young, elderly, pregnant, men, and women}, with proportions shown in Figure~\ref{fig:distribution}(a)(b)(c). Beyond these categories, the dataset includes variations in body shapes, proportions, and poses commonly found in virtual try-on scenarios, including standing, hands-on-hips, raised-arm, walking/running, sitting, and waving poses. A total of 184 human images were collected. Representative examples of the eight clothing categories and nine human categories are presented in Figure~\ref{fig:cloth_type} and Figure~\ref{fig:person_type}. The final paired dataset contains 748 garment–person pairs. Since some algorithms cannot process certain images, manual filtering was applied, resulting in 8,132 generated images from various virtual try-on algorithms.

\subsection{Virtual try-on algorithms}
To generate virtual try-on images, we include a diverse set of representative algorithms spanning traditional and modern paradigms. Specifically, as shown in Table~\ref{tab:vton-algorithms}, our evaluation covers:
\noindent\textbf{(1) Flow- and warp-based two-stage architectures,} representing the \textbf{classical} pipeline of alignment–warping followed by refinement.
\noindent\textbf{(2) Diffusion-based virtual try-on models,} leveraging generative diffusion processes to improve realism and garment fidelity.
\noindent\textbf{(3) Several closed-source commercial or semi-commercial systems,} included to provide references to real-world performance. This comprehensive selection ensures that our dataset and evaluation protocols remain compatible with both earlier pipelines and the latest state-of-the-art virtual try-on techniques.

\begin{table}[t]
\centering
\small
\caption{Categories of virtual try-on algorithms used in our evaluation.}
\vspace{-5pt}
\label{tab:vton-algorithms}
\resizebox{\linewidth}{!}{
\begin{tabular}{lcccc}
\toprule
Model & Year & Resolution  & Type \\
\midrule
VITON-HD \cite{choi2021viton}  & 2021 & 1024x768    & Classical (Warp-based) \\
TPD \cite{Yang_2024_CVPR} & 2024 & 384x512  & Classical (Warp-based) \\
DS-VTON \cite{sun2025dsvtonhighqualityvirtualtryon}  & 2025 & 768×1024    & Classical (Warp-based) \\
FS-VTON \cite{he2022fs_vton}  & 2022 & 256×192  & Classical (Warp-based)\\

\midrule
Ladi-VTON \cite{morelli2023ladi} & 2023 & 1024×768    & Diffusion-based \\
CAT-DM \cite{zeng2024cat} & 2024 & 512×384   & Diffusion-based \\
OOTDiffusion \cite{xu2024ootdiffusion} & 2024 & 1024×768  & Diffusion-based \\
StableVITON \cite{kim2024stableviton} & 2024 & 1024×768   & Diffusion-based \\
CatV2TON \cite{chong2025catv2tontamingdiffusiontransformers} & 2025 & 192×256  & Diffusion-based \\

\midrule
Kling \cite{kling2024tryon}     & 2014 & 512x512-4096x4096   & Closed-source  \\
LinkFox \cite{linkfox2024aidressing}    & 2024 & 384x384-4096x4096  & Closed-source  \\
\bottomrule
\end{tabular}
}

\vspace{-10pt}
\end{table}

\subsection{Subjective Experiment}

After obtaining the virtual try-on results generated by all algorithms, we organized the images into eight non-overlapping groups. A total of 40 independent volunteers with undergraduate and graduate education backgrounds were recruited for subjective evaluation. To facilitate large-scale annotation, each volunteer could participate in the evaluation of one or multiple groups.

\begin{figure}[t]
    \centering
    \includegraphics[width=0.9\linewidth]{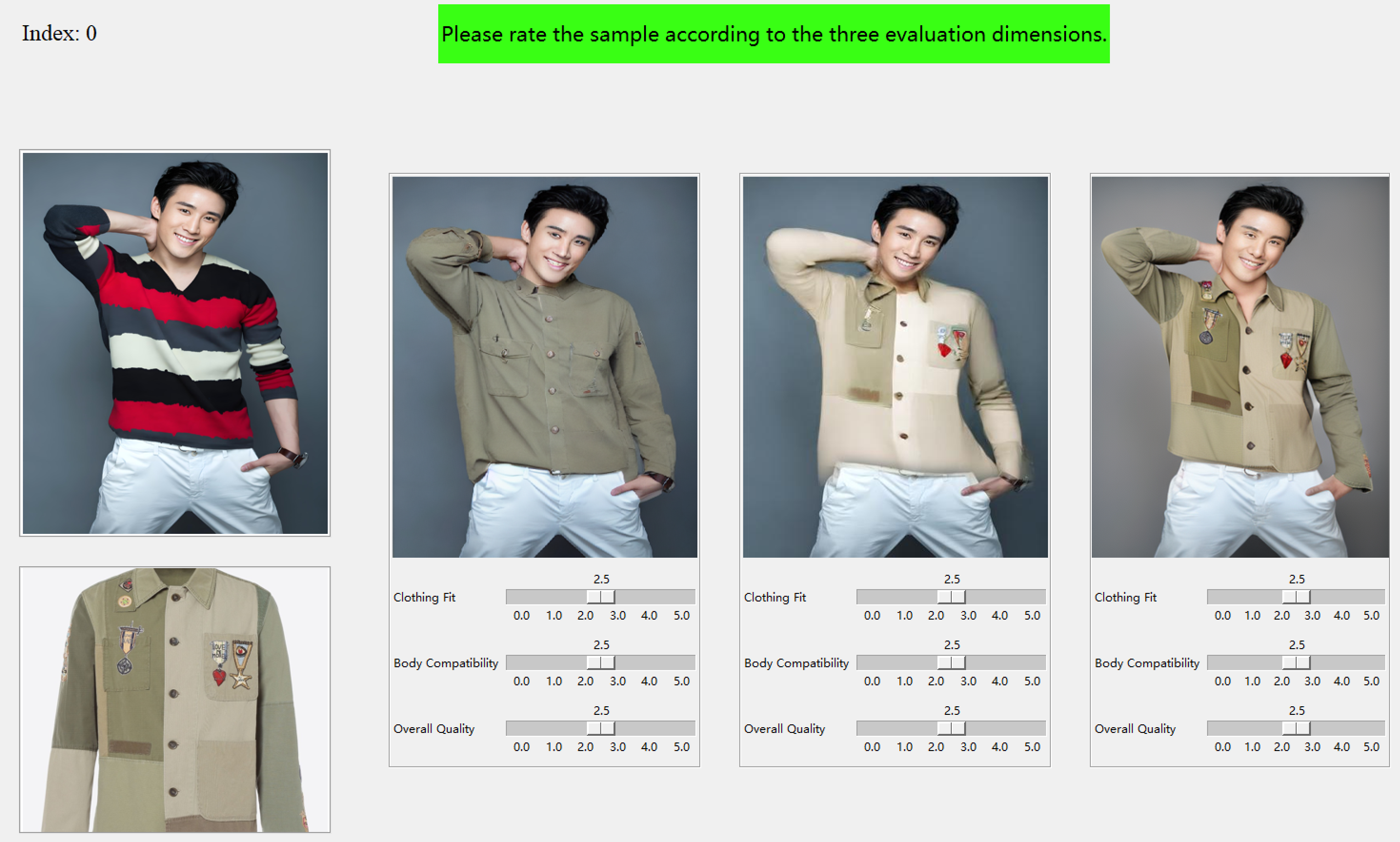}
    \caption{GUI used for subjective quality assessment, where annotators independently rate each virtual try-on result on Clothing Fit, Body Compatibility, and Overall Quality using a 1--5 scale.}
    \vspace{-12pt}
    \label{fig:rate}
\end{figure}

\begin{figure}[tb] \centering
    \includegraphics[width=0.48\textwidth,height=0.10\textwidth]{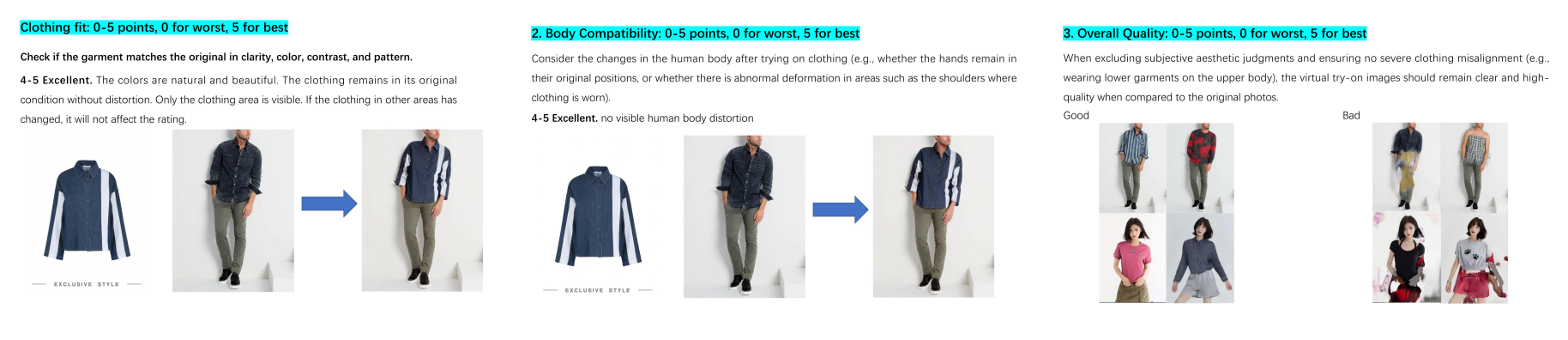}
    \vspace{-18pt}
    \caption{Tutorial presentation for participants in the subjective study} \label{fig:img1}
    \vspace{-1em}
    \label{fig:ppt}
    \vspace{-7pt}
\end{figure}

The evaluation was conducted along three dimensions, which were scored independently on a 1--5 scale with a precision of 0.1:

  \noindent\textbf{(1) Clothing fit:} This metric evaluates whether the target garment is correctly and completely worn in the virtual try-on result.
  \noindent\textbf{(2) Body compatibility:} This metric reflects whether the human body shape and pose remain physically consistent after virtual try-on.
  \noindent\textbf{(3) Overall quality:} This metric measures whether the final synthesized result aligns with human aesthetic perception. Following data collection, all scores were normalized to ensure consistency across evaluators and to facilitate subsequent statistical analysis and comparison with baseline algorithms. The scoring interface is shown in Figure~\ref{fig:rate}. In addition, prior to the subjective experiments, all participants received standardized training to ensure a clear understanding of the three evaluation dimensions, supported by a dedicated presentation (Figure~\ref{fig:ppt}) for detailed explanation.

\subsection{Subjective Data Processing}

We follow the subjective score processing protocol in \cite{bt2002methodology} for outlier detection and subject reliability screening. For each image, a rating is an outlier if it deviates from the mean score by more than $2\sigma$ (normal distributions) or $\sqrt{20}\sigma$ (non-normal distributions). If over $5\%$ of a subject’s ratings are outliers and are approximately symmetrically distributed across score ranges, all ratings from that subject are excluded.
\begin{figure}[t]
    \centering

    \begin{subfigure}[t]{\linewidth}
        \centering
        \begin{tikzpicture}
            \node[inner sep=0pt] (img) at (0,0) {\includegraphics[width=0.18\linewidth,height=2.6cm]{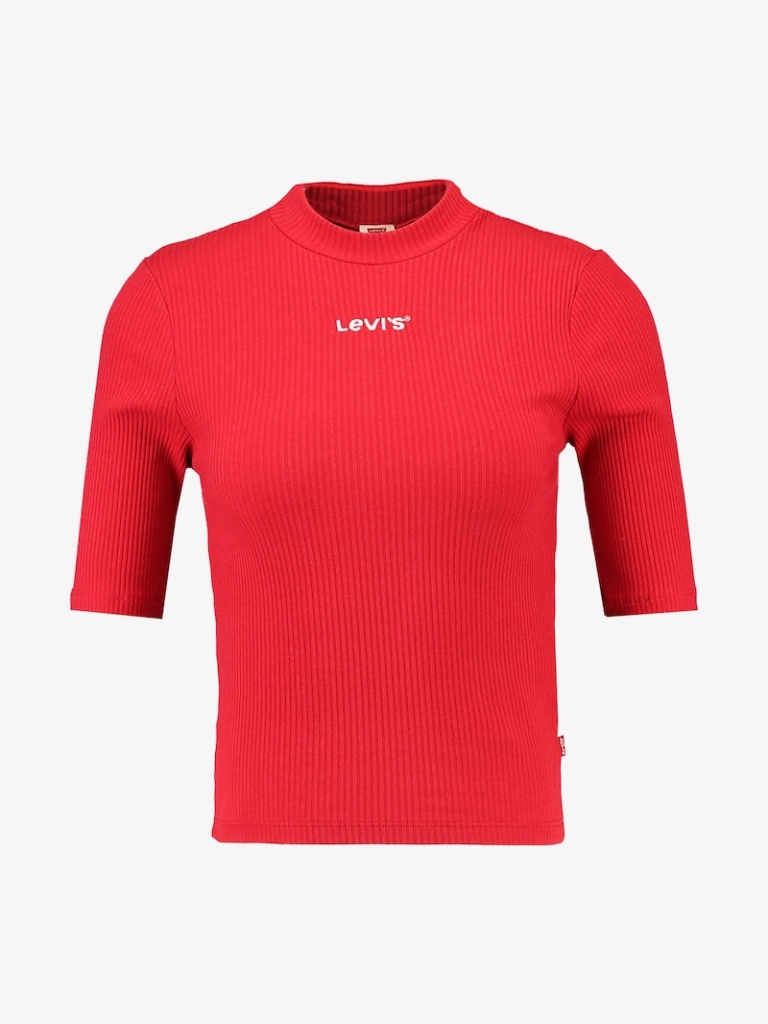}};
            \node[anchor=south east] at (img.south east) [xshift=-2pt,yshift=2pt,fill=black!70,text=white,font=\footnotesize,inner sep=1pt]{\shortstack{Cloth: \\ T-shirt}};
        \end{tikzpicture}
        \begin{tikzpicture}
            \node[inner sep=0pt] (img) at (0,0) {\includegraphics[width=0.18\linewidth,height=2.6cm]{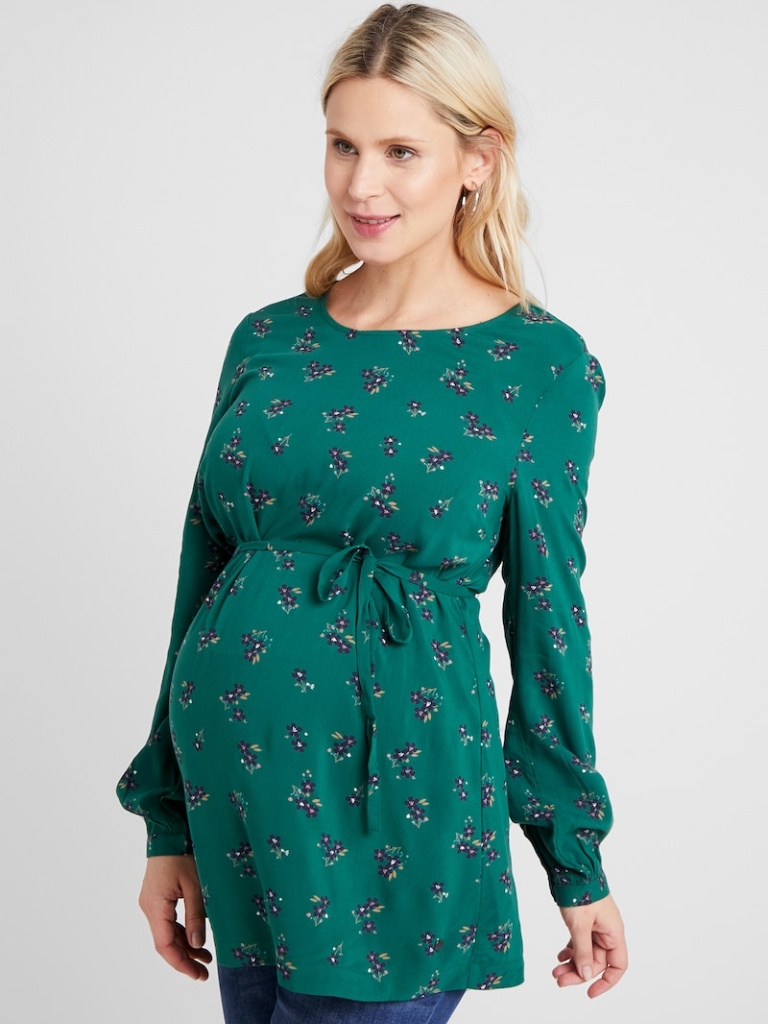}};
            \node[anchor=south east] at (img.south east) [xshift=-2pt,yshift=2pt,fill=black!70,text=white,font=\footnotesize,inner sep=1pt]{\shortstack{Person: \\ pregnent}};
        \end{tikzpicture}
        \begin{tikzpicture}
            \node[inner sep=0pt] (img) at (0,0) {\includegraphics[width=0.18\linewidth,height=2.6cm]{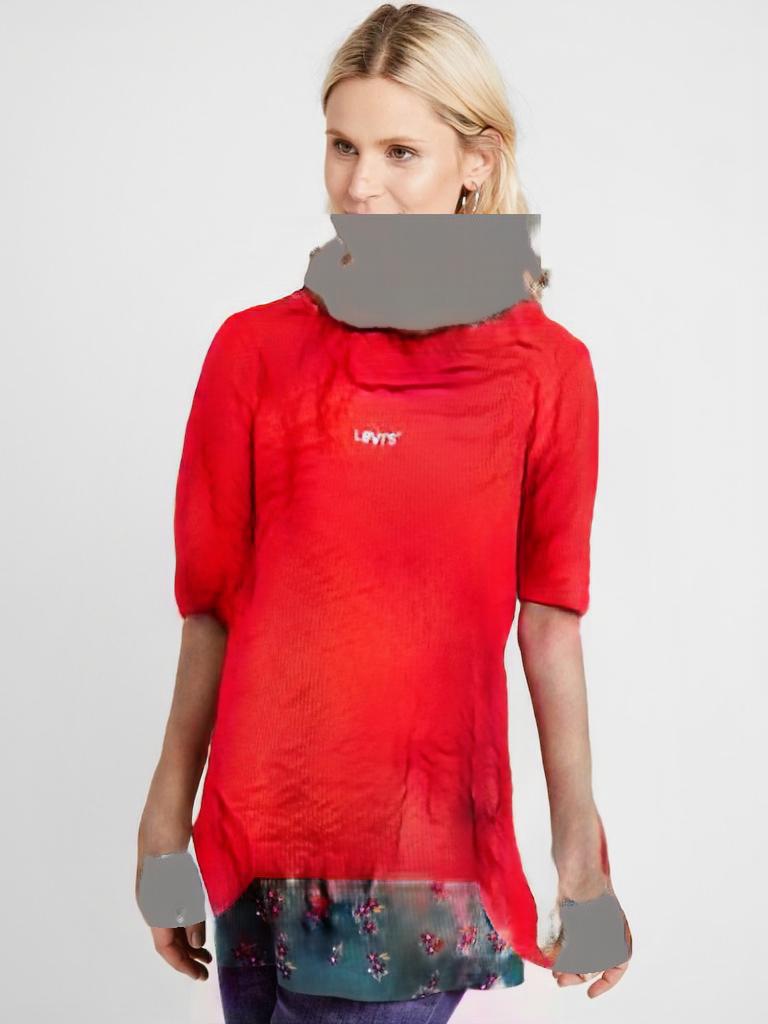}};
            \node[anchor=south east] at (img.south east) [xshift=-2pt,yshift=2pt,fill=black!70,text=white,font=\footnotesize,inner sep=1pt]{37.33};
        \end{tikzpicture}
        \begin{tikzpicture}
            \node[inner sep=0pt] (img) at (0,0) {\includegraphics[width=0.18\linewidth,height=2.6cm]{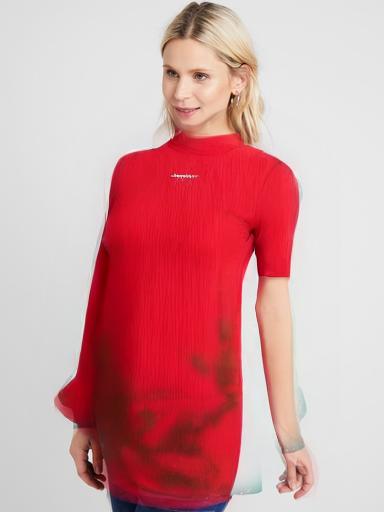}};
            \node[anchor=south east] at (img.south east) [xshift=-2pt,yshift=2pt,fill=black!70,text=white,font=\footnotesize,inner sep=1pt]{54.26};
        \end{tikzpicture}
        \begin{tikzpicture}
            \node[inner sep=0pt] (img) at (0,0) {\includegraphics[width=0.18\linewidth,height=2.6cm]{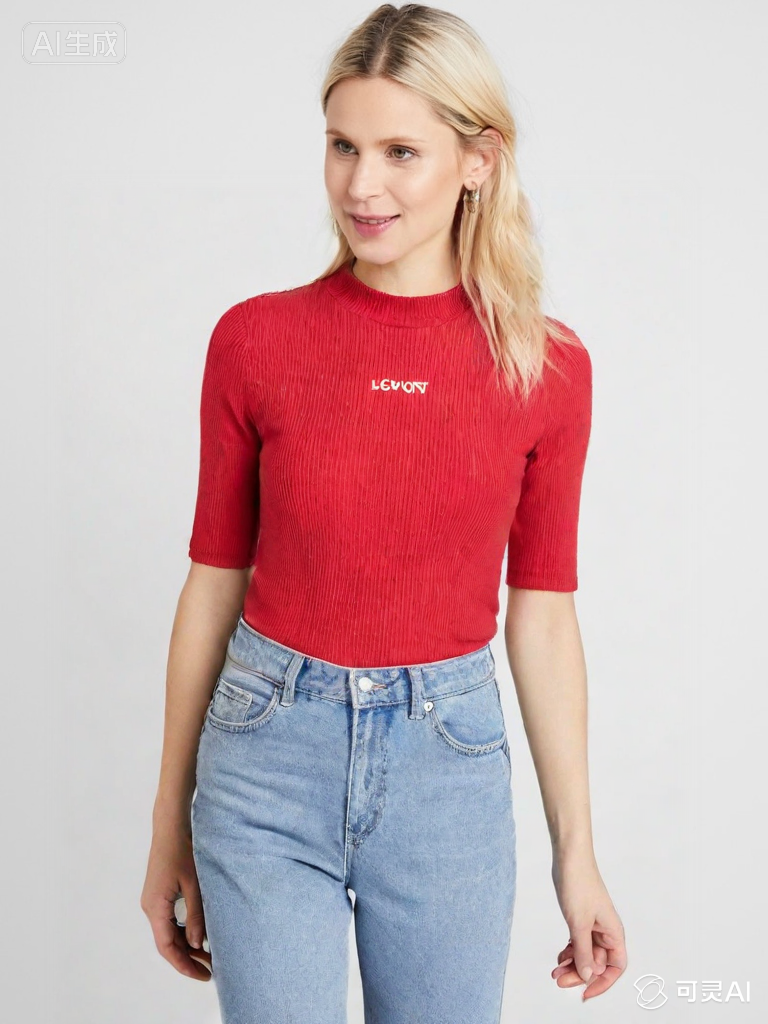}};
            \node[anchor=south east] at (img.south east) [xshift=-2pt,yshift=2pt,fill=black!70,text=white,font=\footnotesize,inner sep=1pt]{68.52};
        \end{tikzpicture}
        \vspace{-3pt}
        \subcaption{Clothing fit}

    \end{subfigure}

    \vspace{6pt}

    \begin{subfigure}[t]{\linewidth}
        \centering
        \begin{tikzpicture}
            \node[inner sep=0pt] (img) at (0,0) {\includegraphics[width=0.18\linewidth,height=2.6cm]{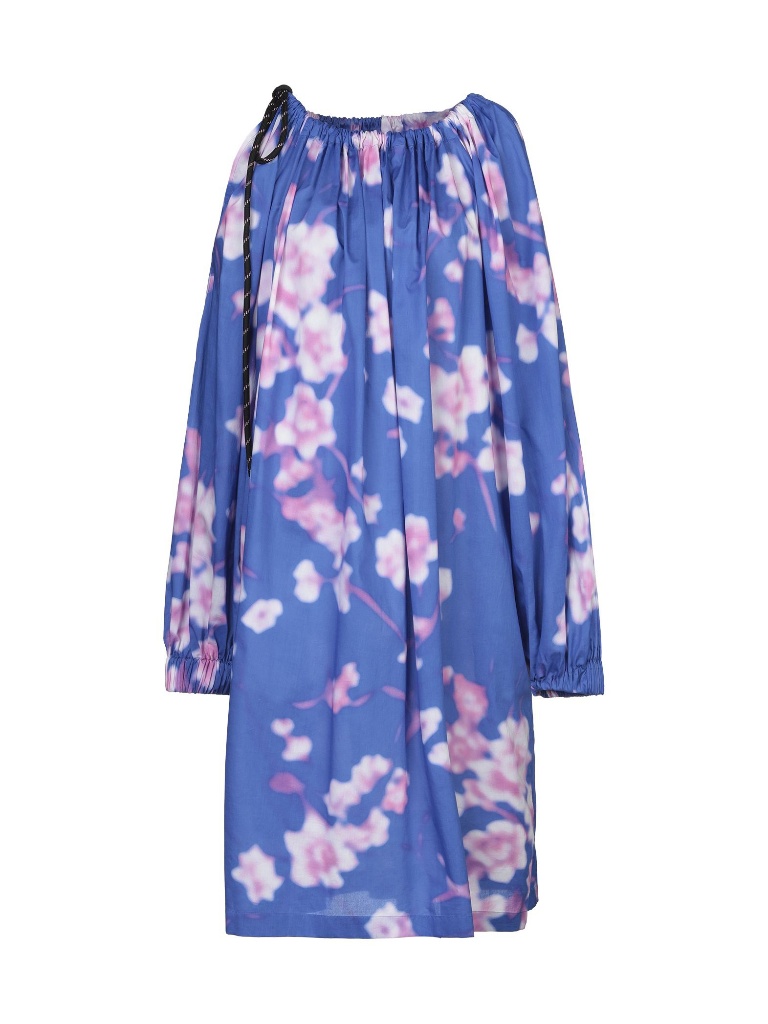}};
            \node[anchor=south east] at (img.south east) [xshift=-2pt,yshift=2pt,fill=black!70,text=white,font=\footnotesize,inner sep=1pt]{\shortstack{Cloth: \\ dress}};
        \end{tikzpicture}
        \begin{tikzpicture}
            \node[inner sep=0pt] (img) at (0,0) {\includegraphics[width=0.18\linewidth,height=2.6cm]{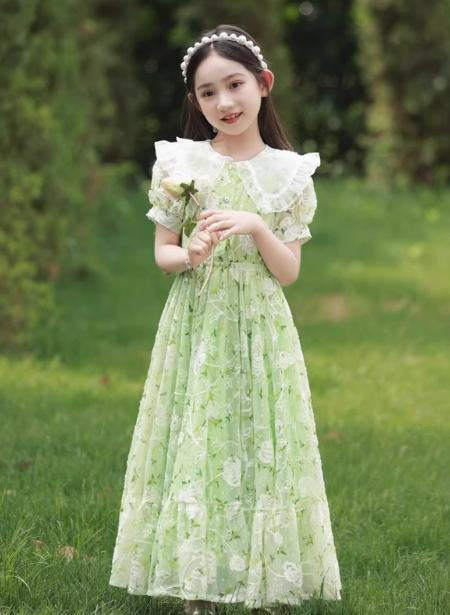}};
            \node[anchor=south east] at (img.south east) [xshift=-2pt,yshift=2pt,fill=black!70,text=white,font=\footnotesize,inner sep=1pt]{\shortstack{Person: \\ children}};
        \end{tikzpicture}
        \begin{tikzpicture}
            \node[inner sep=0pt] (img) at (0,0) {\includegraphics[width=0.18\linewidth,height=2.6cm]{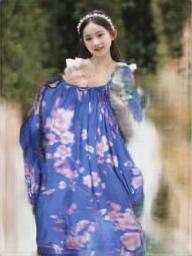}};
            \node[anchor=south east] at (img.south east) [xshift=-2pt,yshift=2pt,fill=black!70,text=white,font=\footnotesize,inner sep=1pt]{22.65};
        \end{tikzpicture}
        \begin{tikzpicture}
            \node[inner sep=0pt] (img) at (0,0) {\includegraphics[width=0.18\linewidth,height=2.6cm]{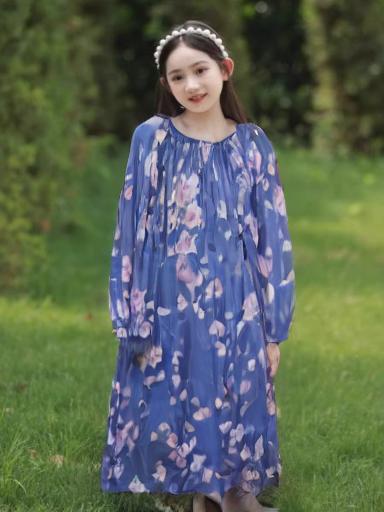}};
            \node[anchor=south east] at (img.south east) [xshift=-2pt,yshift=2pt,fill=black!70,text=white,font=\footnotesize,inner sep=1pt]{48.48};
        \end{tikzpicture}
        \begin{tikzpicture}
            \node[inner sep=0pt] (img) at (0,0) {\includegraphics[width=0.18\linewidth,height=2.6cm]{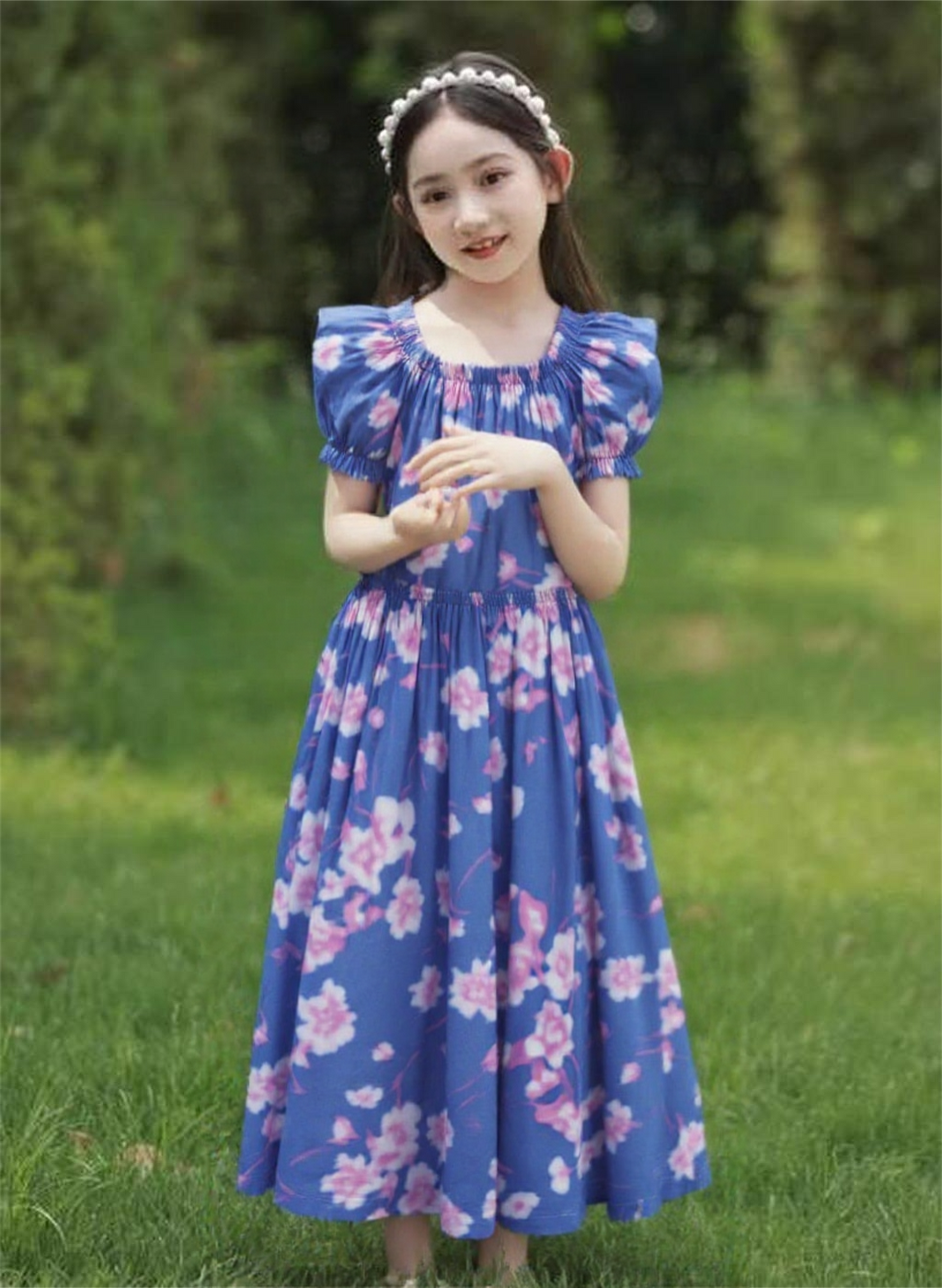}};
            \node[anchor=south east] at (img.south east) [xshift=-2pt,yshift=2pt,fill=black!70,text=white,font=\footnotesize,inner sep=1pt]{63.82};
        \end{tikzpicture}
        \vspace{-3pt}
        \subcaption{Body compatibility}

    \end{subfigure}

    \vspace{6pt}

    \begin{subfigure}[t]{\linewidth}
        \centering
        \begin{tikzpicture}
            \node[inner sep=0pt] (img) at (0,0) {\includegraphics[width=0.18\linewidth,height=2.6cm]{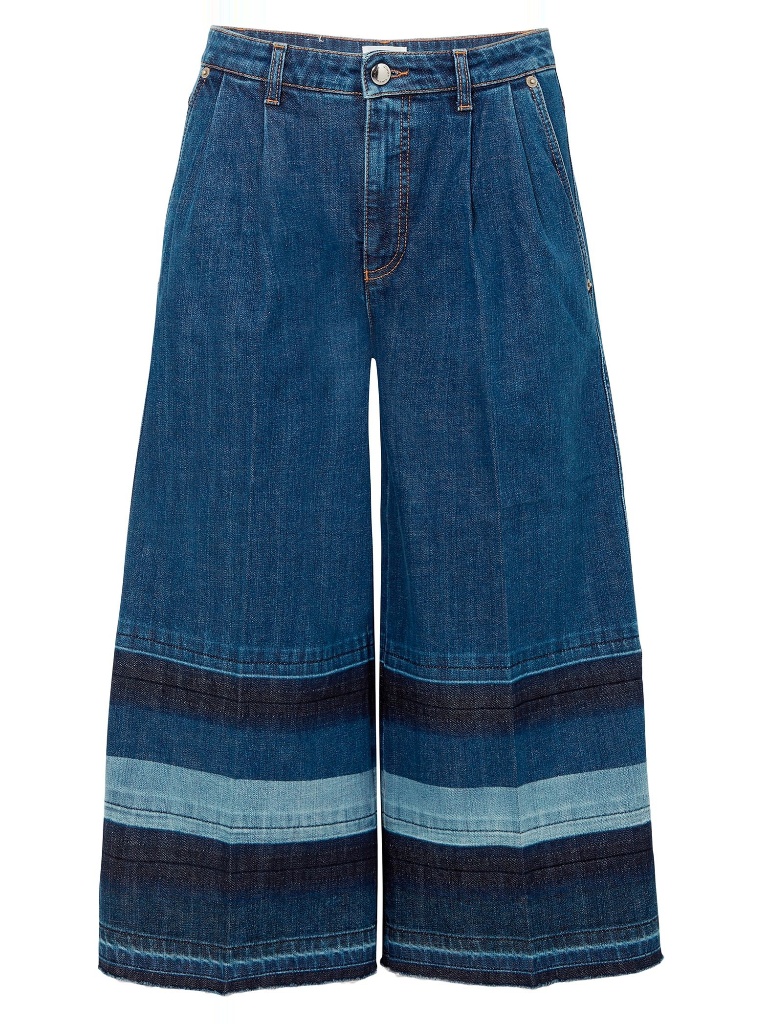}};
            \node[anchor=south east] at (img.south east) [xshift=-2pt,yshift=2pt,fill=black!70,text=white,font=\footnotesize,inner sep=1pt]{\shortstack{Cloth: \\ trousers}};
        \end{tikzpicture}
        \begin{tikzpicture}
            \node[inner sep=0pt] (img) at (0,0) {\includegraphics[width=0.18\linewidth,height=2.6cm]{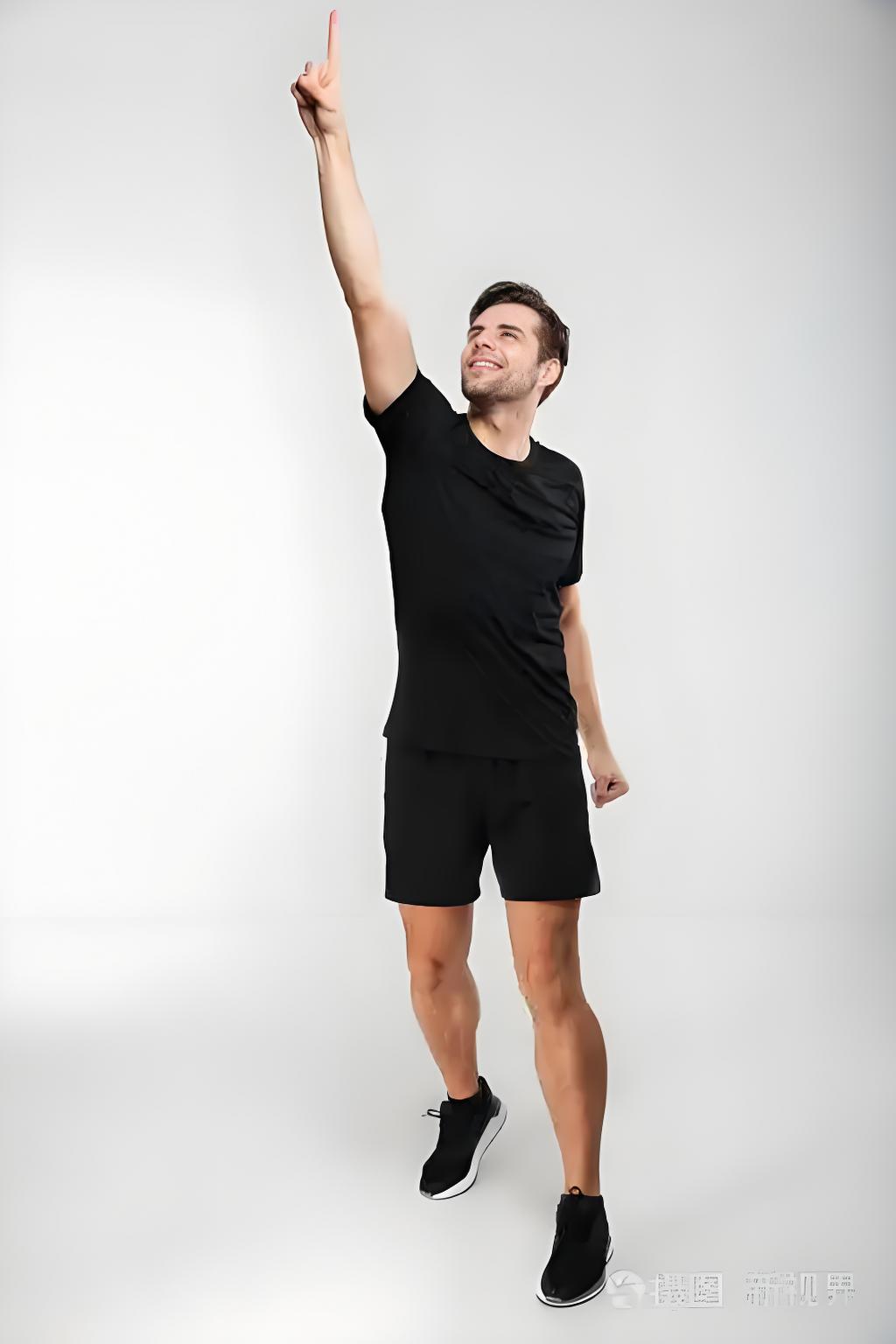}};
            \node[anchor=south east] at (img.south east) [xshift=-2pt,yshift=2pt,fill=black!70,text=white,font=\footnotesize,inner sep=1pt]{\shortstack{Person: \\ man}};
        \end{tikzpicture}
        \begin{tikzpicture}
            \node[inner sep=0pt] (img) at (0,0) {\includegraphics[width=0.18\linewidth,height=2.6cm]{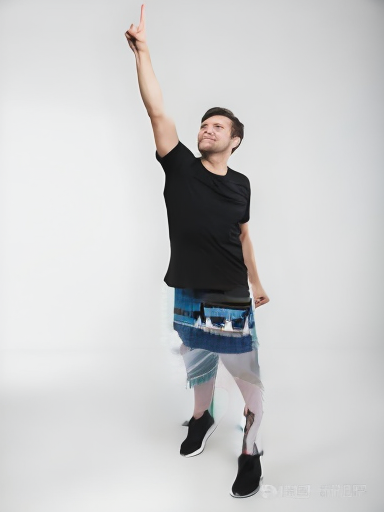}};
            \node[anchor=south east] at (img.south east) [xshift=-2pt,yshift=2pt,fill=black!70,text=white,font=\footnotesize,inner sep=1pt]{33.73};
        \end{tikzpicture}
        \begin{tikzpicture}
            \node[inner sep=0pt] (img) at (0,0) {\includegraphics[width=0.18\linewidth,height=2.6cm]{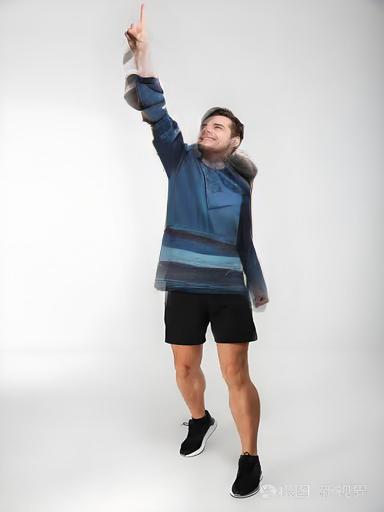}};
            \node[anchor=south east] at (img.south east) [xshift=-2pt,yshift=2pt,fill=black!70,text=white,font=\footnotesize,inner sep=1pt]{41.69};
        \end{tikzpicture}
        \begin{tikzpicture}
            \node[inner sep=0pt] (img) at (0,0) {\includegraphics[width=0.18\linewidth,height=2.6cm]{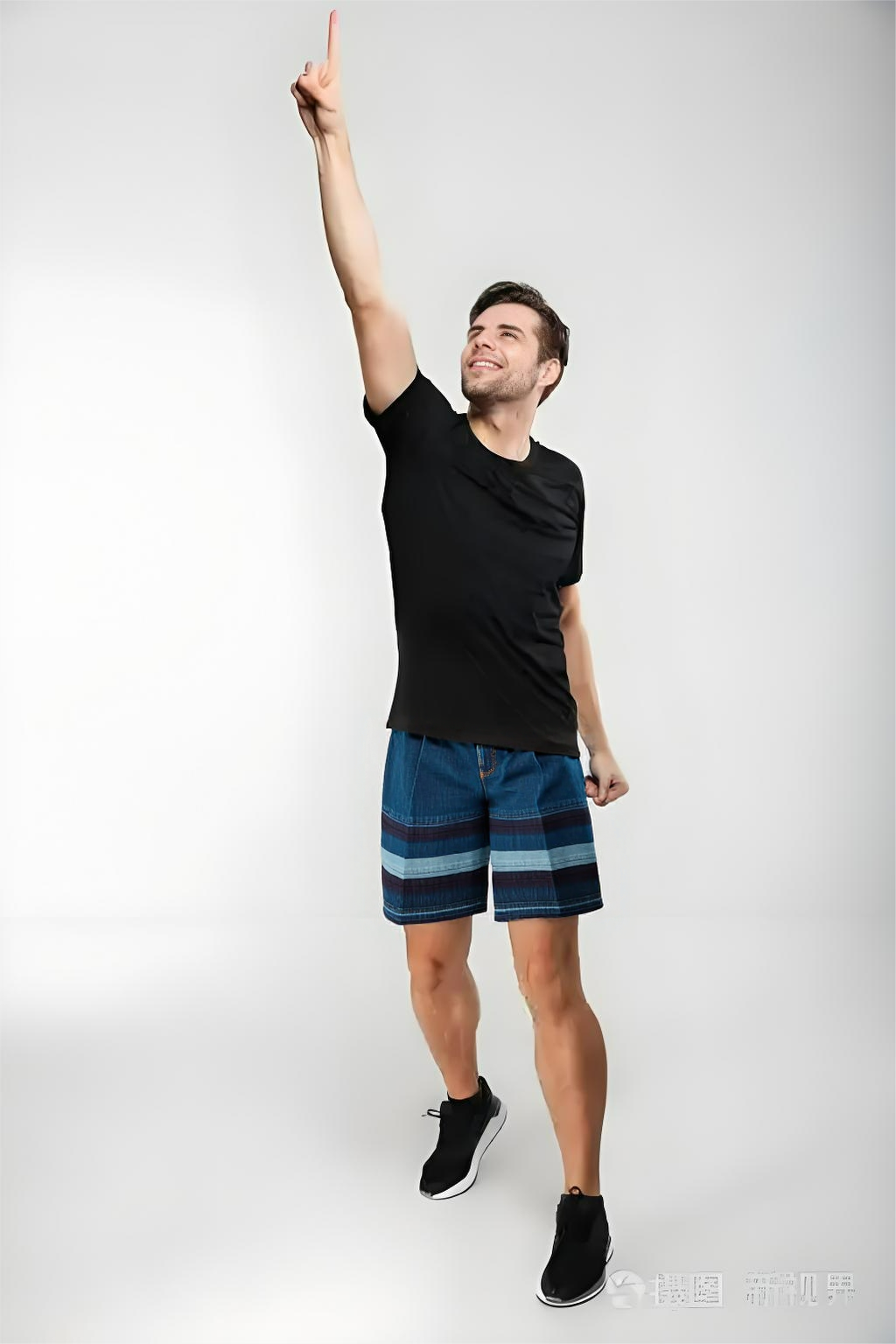}};
            \node[anchor=south east] at (img.south east) [xshift=-2pt,yshift=2pt,fill=black!70,text=white,font=\footnotesize,inner sep=1pt]{65.31};
        \end{tikzpicture}
        \vspace{-3pt}
        \subcaption{Overall quality}

    \end{subfigure}

    \vspace{-8pt}
    \caption{Examples from the proposed VTONQA dataset. We illustrate poor (20–40), average (40–60), and good (60–80) cases for three evaluation dimensions: (a) clothing fit, (b) body compatibility, and (c) overall quality.}
    \vspace{-20pt}
    \label{fig:qualitative_examples}
\end{figure}






\begin{figure*}[t]
    \centering

    \begin{subfigure}{0.31\linewidth}
        \centering
        \includegraphics[width=\linewidth]{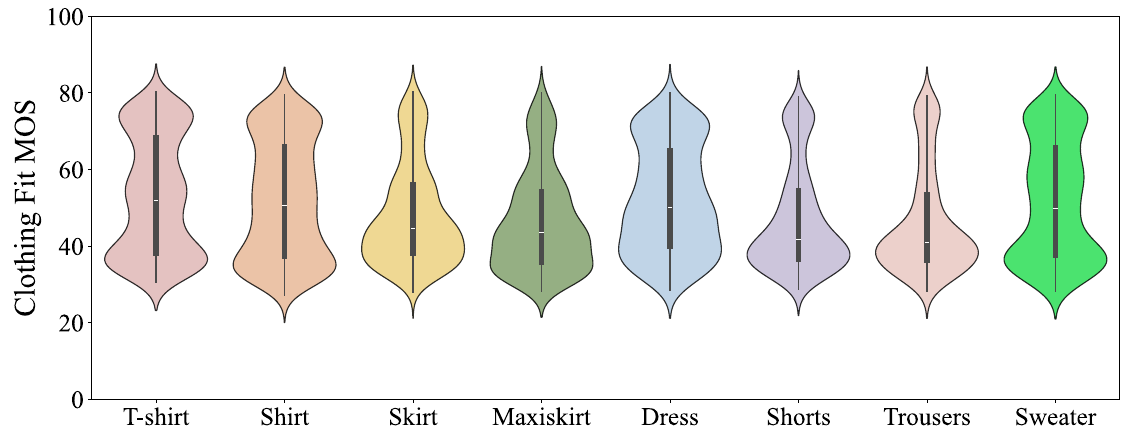}
    \end{subfigure}
    \hfill
    \begin{subfigure}{0.31\linewidth}
        \centering
        \includegraphics[width=\linewidth]{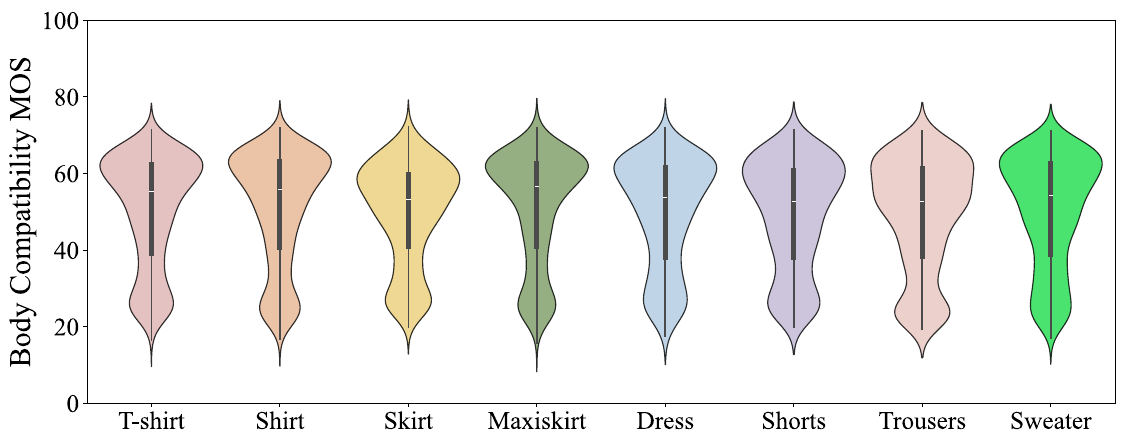}
    \end{subfigure}
    \hfill
    \begin{subfigure}{0.31\linewidth}
        \centering
        \includegraphics[width=\linewidth]{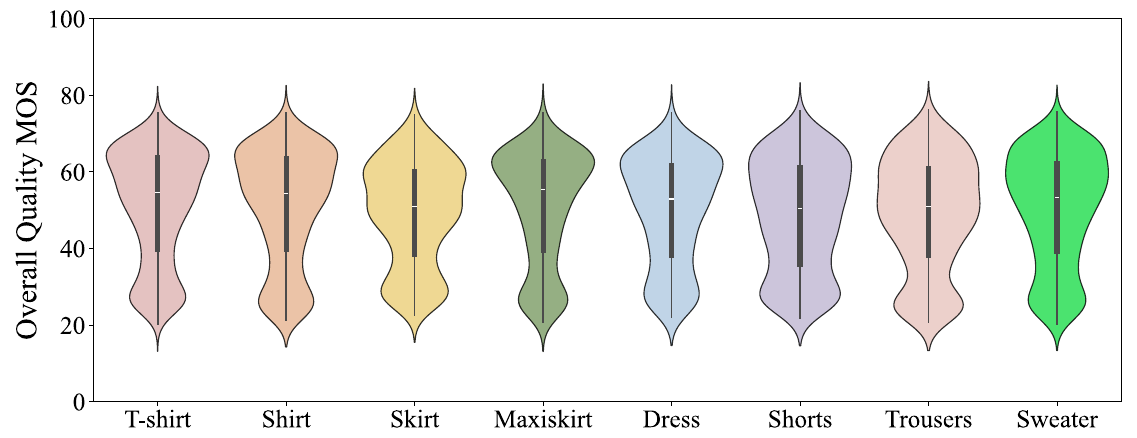}
    \end{subfigure}

    \vspace{6pt}

    \begin{subfigure}{0.31\linewidth}
        \centering
        \includegraphics[width=\linewidth]{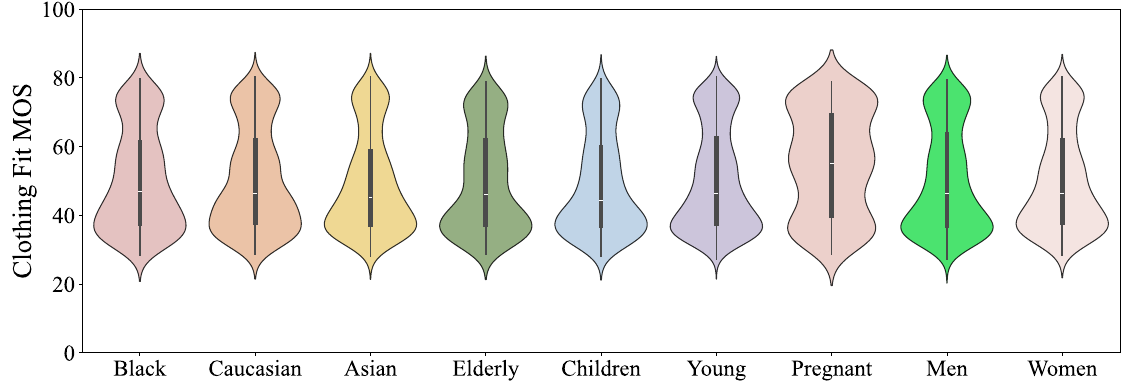}
    \end{subfigure}
    \hfill
    \begin{subfigure}{0.31\linewidth}
        \centering
        \includegraphics[width=\linewidth]{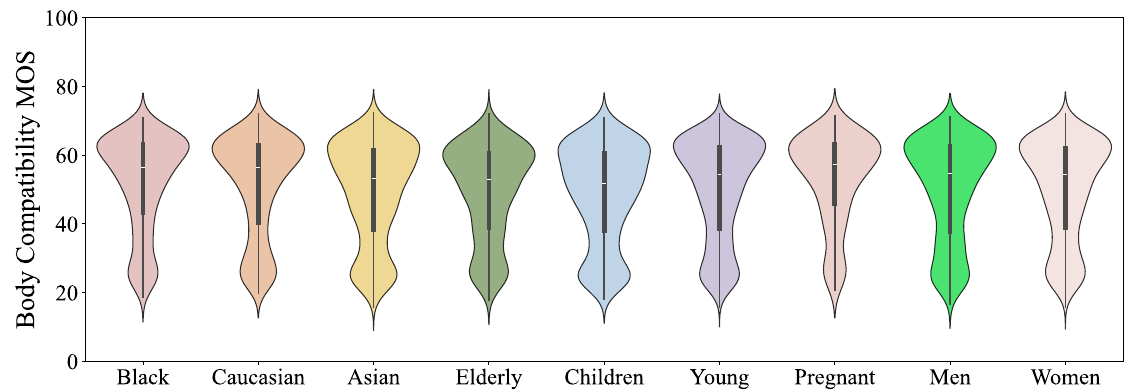}
    \end{subfigure}
    \hfill
    \begin{subfigure}{0.31\linewidth}
        \centering
        \includegraphics[width=\linewidth]{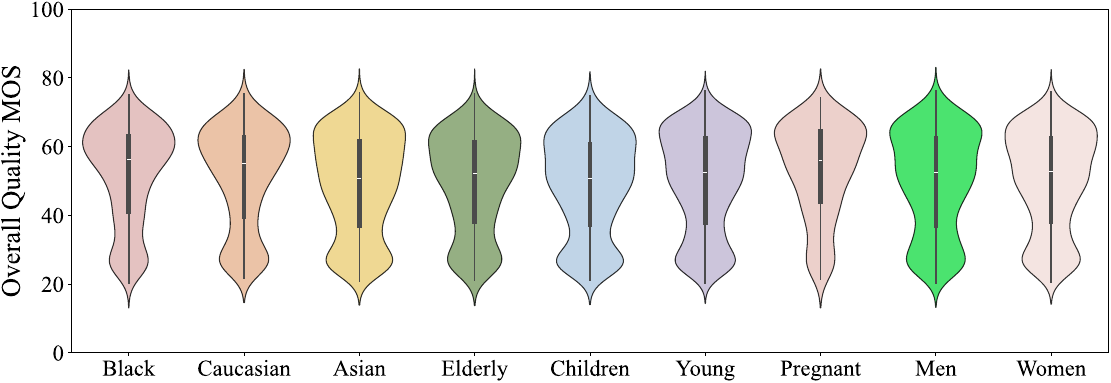}
    \end{subfigure}

    \caption{
    MOS distributions of three evaluation dimensions:
    (a) garment categories and (b) human body categories.
    }

    \label{fig:clothfit_analysis}
    \vspace{-5pt}
\end{figure*}

After filtering, the remaining valid scores are normalized using within-subject Z-score normalization and linearly mapped to the range ($[0, 100]$). The Mean Opinion Score (MOS) for each image is then computed by averaging the normalized scores across all valid subjects, formulated as:

\begin{equation}
\begin{aligned}
z_{ij} &= \frac{r_{ij} - \mu_i}{\sigma_i} ,\quad
s_{ij} = \frac{z_{ij} + 3}{6} \times 100 ,\quad
\mathrm{MOS}_j = \frac{1}{N_j} \sum_{i=1}^{N_j} s_{ij}
\end{aligned}
\end{equation}

where $r_{ij}$ denotes the raw score given by the $i$-th subject to the $j$-th image, $\mu_i$ and $\sigma_i$ represent the mean and standard deviation of all scores provided by subject $i$, respectively, and $N_j$ is the number of valid ratings for image $j$.

To evaluate annotation consistency, we perform a partition-based agreement analysis. Since each annotator scores only part of the dataset, annotators within each group are randomly split into two subsets, from which MOS values are computed. SRCC and PLCC between the two MOS sets are calculated over 10 splits and averaged. The results show agreement across all dimensions: Clothing Fit (SRCC/PLCC: $0.7456 \pm 0.0253$ / $0.7774 \pm 0.0195$), Body Compatibility ($0.7348 \pm 0.0295$ / $0.7811 \pm 0.0341$), and Overall Quality ($0.7572 \pm 0.0273$ / $0.7913 \pm 0.0306$), demonstrating annotation consistency.

\subsection{Multi-dimensional Analysis}
\begin{figure}[t]
    \centering
    \includegraphics[width=0.85\linewidth]{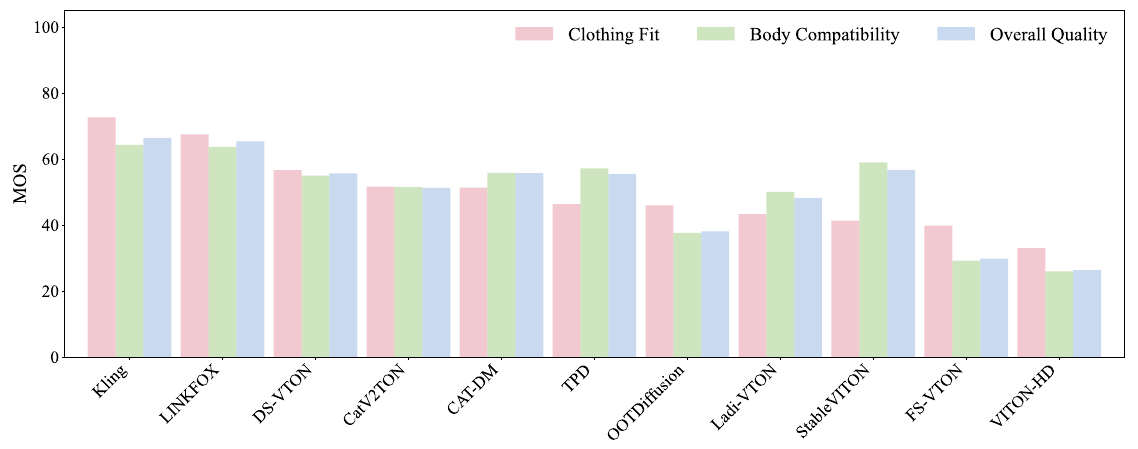}
    \vspace{-8pt}
    \caption{Comparison of the 11 VTON models based on average clothing fit, body compatibility, and overall quality scores.}
    \vspace{-16pt}
    \label{fig:all_model}
\end{figure}

We present a three-dimensional evaluation demo with the corresponding MOS results in Figure \ref{fig:qualitative_examples}. The three illustrated score ranges, ordered from low to high, reflect performance across different evaluation dimensions: scores below 40 are categorized as poor, scores between 40 and 60 as average, and scores above 60 as good. As shown in Figure~\ref{fig:all_model}, performance across the three evaluation dimensions shows consistent trends. Overall, the two closed-source systems outperform all open-source virtual try-on models by a clear margin. Among the remaining algorithms, DS-VTON achieves the best performance within the classical (warp-based) category, while StableVITON leads the diffusion-based methods. Notably, diffusion-based approaches exhibit stronger robustness across diverse scenarios. Although each class of algorithms demonstrates strengths on specific dimensions, performance gaps among open-source methods remain relatively moderate.

\begin{figure}[t]
    \centering
    \includegraphics[width=0.9\linewidth]{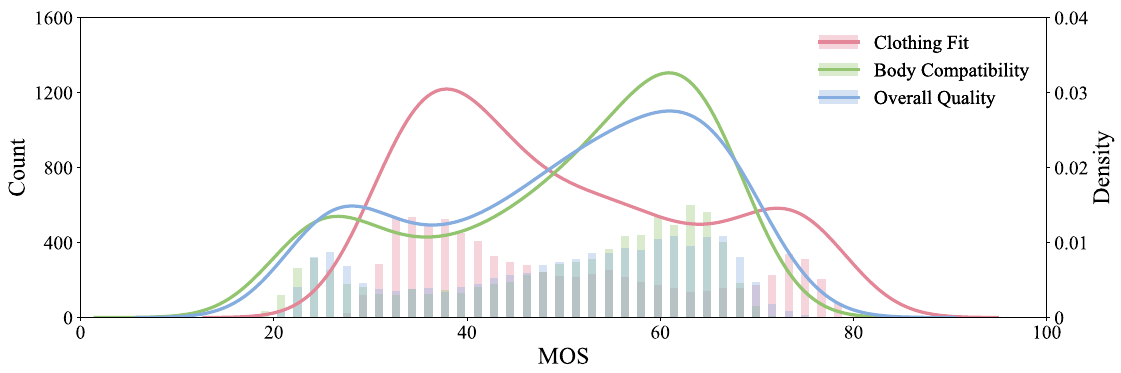}
    \vspace{-5pt}
    \caption{MOS distribution histograms and kernel density curves for clothing fit, body compatibility, and overall quality.}
    \vspace{-18pt}
    \label{fig:all_mos}
\end{figure}

The overall distribution of subjective scores is shown in Figure~\ref{fig:all_mos}. The scores mainly fall within the range of 20–80, where values between 60–80 indicate strong performance and those between 20–40 reflect weaker results. As illustrated by the distribution, most virtual try-on algorithms achieve relatively high scores in body compatibility, suggesting that current models generally preserve human pose without introducing significant structural distortions. In contrast, far fewer methods perform well in clothing fit, that is, accurately aligning the target clothing with the person. This highlights a substantial performance gap when algorithms operate under complex or realistic scenarios. Furthermore, the distribution indicates that overall quality is more heavily influenced by body compatibility than by clothing fit alone.

\begin{table*}[t]
\centering
\scriptsize
\setlength{\tabcolsep}{3pt} 
\caption{Evaluation of 11 representative VTON models based on the overall quality score. We report both the overall average score and the scores across 8 garment categories and 9 human body categories. $\spadesuit$ classical (warp-based) method, $\heartsuit$ diffusion-based method, and $\clubsuit$ closed-source method. The best results are highlighted in \textcolor{red}{red}, the second-best results are highlighted in \textcolor{blue}{blue}, the third-best results are highlighted in \textcolor{thirdcolor}{green}.}
\vspace{-5pt}
\label{tab:clothing_fit_mos_category_attribute}

\begin{tabular}{lcccccccccccccccccc}
\toprule
\multirow{2}{*}{\textbf{Methods}} &
\multicolumn{8}{c}{\textbf{Clothing Categories}} &
\multicolumn{9}{c}{\textbf{Human Body Categories}} &
\multirow{2}{*}{\textbf{Overall}} \\
\cmidrule(lr){2-9} \cmidrule(lr){10-18}
 & \textbf{T-shirt} & \textbf{Shirt} & \textbf{Sweater} & \textbf{Shorts} & \textbf{Trousers} & \textbf{Maxiskirt} & \textbf{Skirt} & \textbf{Dress}
 & \textbf{Black} & \textbf{Caucasian} & \textbf{Asian} & \textbf{Children} & \textbf{Young} & \textbf{Elderly} & \textbf{Pregnant} & \textbf{Man} & \textbf{Woman} & \\
\midrule

$\spadesuit$VITON-HD \cite{choi2021viton} 
 & 27.21 & 25.78 & 25.77 & 26.58 & 25.07 & 26.28 & 27.73 & 25.25
 & 26.22 & 26.22 & 26.28 & 25.80 & 26.62 & 26.12 & 25.73 & 26.71 & 26.71 & 26.26\\

$\spadesuit$TPD \cite{Yang_2024_CVPR}
 & 56.89 & 54.56 & 53.78 & 55.11 & \textcolor{thirdcolor}{56.85} & 56.62 & 53.66 & \textcolor{thirdcolor}{56.95}
 & 56.10 & 56.09 & 54.87 & 54.74 & 55.55 & 54.16 & 56.84 & 55.91 & 55.28 & 55.48 \\

$\spadesuit$DS-VTON \cite{sun2025dsvtonhighqualityvirtualtryon} 
 & \textcolor{thirdcolor}{60.39} & \textcolor{thirdcolor}{61.15} & \textcolor{thirdcolor}{59.28} & 49.66 & 49.57 & 53.86 & 51.04 & 54.01
 & 56.19 & 56.60 & 55.37 & 52.83 & 55.24 & 54.21 & \textcolor{thirdcolor}{57.64} & \textcolor{thirdcolor}{56.23} & 54.72 & 55.64 \\

$\spadesuit$FS-VTON \cite{he2022fs_vton}
 & 32.15 & 30.60 & 31.58 & 28.43 & 25.89 & 28.17 & 30.57 & 28.96
 & 31.09 & 30.66 & 29.34 & 27.66 & 29.49 & 29.87 & 35.39 & 28.35 & 29.89 & 29.79 \\
\midrule

$\heartsuit$Ladi-VTON \cite{morelli2023ladi}
 & 49.81 & 49.17 & 47.28 & 44.80 & 47.60 & 52.69 & 47.45 & 46.62
 & 47.10 & 48.89 & 49.28 & 45.33 & 49.07 & 44.50 & 48.42 & 49.05 & 49.01 & 48.21 \\

$\heartsuit$CAT-DM \cite{zeng2024cat} 
 & 56.03 & 56.67 & 55.03 & 54.86 & 54.83 & 59.73 & 55.08 & 52.82
 & \textcolor{thirdcolor}{58.30} & 56.30 & 56.11 & 52.83 & 54.42 & 56.34 & 55.57 & 53.96 & 54.25 & 55.65 \\

$\heartsuit$OOTDiffusion \cite{xu2024ootdiffusion} 
 & 37.63 & 39.16 & 38.90 & 36.64 & 40.38 & 38.99 & 38.88 & 32.94
 & 39.58 & 38.32 & 37.31 & 37.43 & 37.79 & 38.70 & 39.21 & 37.70 & 38.02 & 38.07 \\

$\heartsuit$StableVITON \cite{kim2024stableviton} 
 & 56.22 & 56.21 & 55.44 & \textcolor{thirdcolor}{58.93} & 56.54 & \textcolor{thirdcolor}{60.18} & \textcolor{thirdcolor}{59.32} & 52.72
 & 58.00 & \textcolor{thirdcolor}{56.91} & \textcolor{thirdcolor}{57.17} & \textcolor{thirdcolor}{57.14} & \textcolor{thirdcolor}{55.85} & \textcolor{thirdcolor}{56.59} & 55.67 & 55.41 & \textcolor{thirdcolor}{56.10} & \textcolor{thirdcolor}{56.68} \\

$\heartsuit$CatV2TON \cite{chong2025catv2tontamingdiffusiontransformers}
 & 54.18 & 53.59 & 53.48 & 47.27 & 46.84 & 51.15 & 47.86 & 52.15
 & 51.80 & 51.43 & 51.06 & 49.07 & 51.10 & 50.60 & 52.46 & 50.95 & 51.14 & 51.21 \\
\midrule

$\clubsuit$Kling \cite{kling2024tryon}
 & \textcolor{red}{66.46} & \textcolor{red}{67.00} & \textcolor{red}{67.87} & \textcolor{blue}{65.40} & \textcolor{red}{66.34} & \textcolor{red}{65.50} & \textcolor{red}{64.86} & \textcolor{red}{66.16}
 & \textcolor{red}{66.00} & \textcolor{red}{66.49} & \textcolor{red}{66.50} & \textcolor{red}{65.90} & \textcolor{red}{66.40} & \textcolor{red}{64.77} & \textcolor{red}{65.36} & \textcolor{red}{67.13} & \textcolor{red}{66.15} & \textcolor{red}{66.35} \\

$\clubsuit$LinkFox \cite{linkfox2024aidressing}
 & \textcolor{blue}{65.56} & \textcolor{blue}{65.98} & \textcolor{blue}{64.32} & \textcolor{red}{66.47} & \textcolor{blue}{66.66} & \textcolor{blue}{64.47} & \textcolor{blue}{64.17} & \textcolor{blue}{64.17}
 & \textcolor{blue}{64.29} & \textcolor{blue}{65.02} & \textcolor{blue}{65.55} & \textcolor{blue}{64.34} & \textcolor{blue}{65.69} & \textcolor{blue}{64.19} & \textcolor{blue}{64.57} & \textcolor{blue}{66.04} & \textcolor{blue}{65.62} & \textcolor{blue}{65.28} \\

\bottomrule
\end{tabular}

\vspace{-5pt}
\end{table*}

Beyond the overall evaluation, we further examine garment compatibility as the primary dimension of interest, evaluating algorithm performance from two perspectives: score variations across garment categories and across human body categories. The main observations are summarized as follows:

\noindent\textbf{Garment compatibility.} As shown in Fig.~\ref{fig:clothfit_analysis}(a), virtual try-on performance for upper-body and full-body garments is consistently and noticeably superior to that for lower-body garments across scenarios. This performance gap is substantial, clearly indicating that current algorithms handle upper-body contours and global garment structures more reliably, while remaining significantly more sensitive to deformation-prone lower-body garments.

\noindent\textbf{Performance across human body categories.} As shown in Fig.~\ref{fig:clothfit_analysis}(b), score distributions across human body categories show minor variations, indicating strong generalization. Notably, the pregnant category has a slightly higher proportion of high scores, likely due to dataset construction, as most cases involve upper-body garments where algorithms perform more reliably.


\begin{table*}[t]
\centering
\scriptsize
\setlength{\tabcolsep}{7pt}        
\renewcommand{\arraystretch}{0.7} 

\caption{Comparison of IQA metrics on the VTONQA dataset for predicting clothing fit, body compatibility, and overall quality scores. SRCC, KRCC, and PLCC are reported. $\spadesuit$ traditional full-reference IQA metrics, $\heartsuit$ traditional no-reference IQA metrics, $\clubsuit$ deep learning--based full-reference IQA methods, and $\diamondsuit$ deep learning--based no-reference IQA methods. Fine-tuned results are marked with *. The best results are highlighted in \textcolor{red}{red}, and the second-best results are highlighted in \textcolor{blue}{blue}.}
\vspace{-10pt}
\label{tab:metrics_comparison}

\begin{adjustbox}{width=\textwidth}
\begin{tabular}{lccccccccc}
\toprule
\multirow{2}{*}{\textbf{Methods}} &
\multicolumn{3}{c}{\textbf{Clothing Fit}} &
\multicolumn{3}{c}{\textbf{Body Compatibility}} &
\multicolumn{3}{c}{\textbf{Overall Quality}} \\
\cmidrule(lr){2-4} \cmidrule(lr){5-7} \cmidrule(lr){8-10}
& \textbf{SRCC}$\uparrow$ & \textbf{KRCC}$\uparrow$ & \textbf{PLCC}$\uparrow$
& \textbf{SRCC}$\uparrow$ & \textbf{KRCC}$\uparrow$ & \textbf{PLCC}$\uparrow$
& \textbf{SRCC}$\uparrow$ & \textbf{KRCC}$\uparrow$ & \textbf{PLCC}$\uparrow$\\
\midrule
$\spadesuit$MSE          
& -0.0350 & -0.0258 & -0.0289 
& 0.3117 & -0.2189 & -0.2911 
& 0.2688 & 0.1886 & 0.2483 \\
$\spadesuit$PSNR         
& -0.0350 & -0.0258 & -0.1009 
& 0.3117 & 0.2189 & 0.3592 
& 0.2688 & 0.1886 & 0.3126 \\
$\spadesuit$SSIM \cite{SSIM}  
& 0.0563 & 0.0376 & 0.0794 
& 0.3299 & 0.2250 & 0.3302 
& 0.2912 & 0.1982 & 0.2946 \\
$\spadesuit$FSIM \cite{fsim}  
& 0.0799 & 0.0516 & 0.0478 
& 0.4079 & 0.2848 & 0.4390 
& 0.3743 & 0.2610 & 0.3991 \\
$\spadesuit$SCSSIM \cite{scssim} 
& 0.0394 & 0.0270 & 0.0667 
& 0.3164 & 0.2159 & 0.3147 
& 0.2767 & 0.1888 & 0.2785 \\
$\spadesuit$GMSD \cite{gmsd}   
& 0.1083 & 0.0721 & 0.1168 
& 0.1970 & 0.1325 & 0.1917 
& 0.1947 & 0.1315 & 0.1986 \\
\midrule
$\heartsuit$BRISQUE \cite{brisque}   
& 0.1009 & 0.0654 & 0.1428 
& 0.1778 & 0.1181 & 0.1679 
& 0.1732 & 0.1151 & 0.1735 \\
\midrule
$\clubsuit$LPIPS(alex) \cite{lpips}   
& 0.0615 & 0.0360 & 0.0831 
& 0.4292 & 0.3024 & 0.4967 
& 0.3918 & 0.2761 & 0.4532 \\
$\clubsuit$LPIPS(vgg) \cite{lpips}   
& 0.1395 & 0.0872 & 0.1920 
& \textcolor{blue}{0.4927} & \textcolor{blue}{0.3473} & 0.5522 
& \textcolor{blue}{0.4572} & \textcolor{blue}{0.3214} & \textcolor{blue}{0.5160} \\
$\clubsuit$AHIQ \cite{AHIQ}   
& -0.0716 & -0.0505 & -0.0676 
& 0.2173 & 0.1478 & 0.2605 
& 0.1773 & 0.1204 & 0.2147 \\
\midrule
$\diamondsuit$CNNIQA \cite{CNNIQA}   
& 0.0326 & 0.0231 & -0.1102 
& 0.1126 & 0.0793 & 0.0703 
& 0.1128 & 0.0801 & 0.0446 \\
$\diamondsuit$WaDIQaM \cite{WADIQAM}   
& -0.0030 & 0.0011 & -0.1399
& 0.0872 & 0.0656 & 0.0813 
& 0.0789 & 0.0606 & 0.0493 \\
$\diamondsuit$NIMA \cite{NIMA}   
& 0.3194 & 0.2159 & 0.2969 
& 0.4322 & 0.2966 & \textcolor{blue}{0.5092} 
& 0.4318 & 0.2953 & 0.4969 \\
$\diamondsuit$HyperIQA \cite{HyperIQA}   
& 0.1335 & 0.0759 & 0.1120 
& 0.2876 & 0.1834 & 0.4305 
& 0.2785 & 0.1762 & 0.3984 \\
$\diamondsuit$TOPIQ* \cite{TOPIQ}   
& 0.3157 & 0.2156 & 0.3010 
& 0.3666 & 0.2767 & 0.4586 
& 0.3930 & 0.2745 & 0.4466 \\
$\diamondsuit$MANIQA* \cite{MANIQA}   
& \textcolor{red}{0.6733} & \textcolor{red}{0.4813} & \textcolor{red}{0.6334} 
& \textcolor{red}{0.6654} & \textcolor{red}{0.4813} & \textcolor{red}{0.8011} 
& \textcolor{red}{0.7072} & \textcolor{red}{0.5119} & \textcolor{red}{0.7967} \\
$\diamondsuit$CLIPIQA* \cite{CLIPIQA}   
& \textcolor{blue}{0.4420} & \textcolor{blue}{0.3108} & \textcolor{blue}{0.4190} 
& 0.3010 & 0.2176 & 0.4550 
& 0.3720 & 0.2658 & 0.5000 \\
\bottomrule
\end{tabular}
\end{adjustbox}

\vspace{-6pt}
\end{table*}

\section{Analysis}

\subsection{Basic Analysis}

Table~\ref{tab:clothing_fit_mos_category_attribute} presents processed scores for the Clothing Fit dimension across different garment and human body categories. Beyond the overall comparison, several observations can be drawn. First, diffusion-based methods (e.g., StableVITON) generally outperform classical warp-based approaches (e.g., FS-VTON), while closed-source models consistently achieve the best and most stable performance across categories. Second, garment type plays a more critical role than body category, with lower-body garments (e.g., trousers and shorts) being significantly more challenging than upper-body garments (e.g., T-shirt and shirt). Third, some open-source methods (e.g., DS-VTON and StableVITON) demonstrate competitive performance in specific categories but lack consistency across all settings, indicating a trade-off between specialization and generalization. Finally, the large performance gap among methods highlights the substantial room for improvement in current virtual try-on approaches.

\vspace{-3pt}
\subsection{Baseline Experiment}
We evaluate a diverse set of baseline methods, including: (1) traditional full-reference (FR) IQA metrics: MSE, PSNR, SSIM \cite{SSIM}, FSIM \cite{fsim}, SCSSIM \cite{scssim}, GMSD \cite{gmsd}; (2) traditional no-reference (NR) IQA metrics: BRISQUE \cite{brisque}, (3) deep learning–based FR IQA methods: LPIPS(alex) \cite{lpips}, LPIPS(vgg) \cite{lpips}, AHIQ \cite{AHIQ}; (4) deep learning–based NR IQA methods: CNNIQA \cite{CNNIQA}, WaDIQaM \cite{WADIQAM}, NIMA \cite{NIMA}, HyperIQA \cite{HyperIQA}, TOPIQ \cite{TOPIQ}, MANIQA \cite{MANIQA},  CLIPIQA \cite{CLIPIQA}. Among the deep learning–based NR IQA approaches, several models are further fine-tuned using the proposed subjective dataset. For models without additional training, the virtual try-on images are directly fed into the baseline algorithms to assess which of the three perceptual dimensions the models naturally align with. For models that are further fine-tuned, each dimension’s scores are individually used to train the model, verifying how our dataset can improve dimension-specific evaluation performance. 

We use Spearman’s Rank Correlation Coefficient(SRCC), Pearson Linear Correlation Coefficient(PLCC), and Kendall’s Rank Correlation Coefficient(KRCC) to evaluate the consistency between predicted quality scores and ground-truth MOSs. The resulting values are reported in Table~\ref{tab:metrics_comparison}. 

From the table, several observations emerge: First, traditional IQA metrics—both full-reference (FR) and no-reference (NR)—correlate poorly with human perception of virtual try-on results. Pixel-level similarity metrics are inconsistent, failing to capture garment deformation and body–cloth interactions, while conventional NR-IQA methods based on natural image statistics are limited in this scenario. Second, perceptual distance–based metrics perform better on body compatibility and overall quality but struggle with clothing fit. Third, deep learning–based NR-IQA methods outperform traditional metrics, with transformer- or multi-modal approaches (e.g., MANIQA) achieving the best results. Notably, clothing fit remains the most challenging dimension.

\vspace{-8pt}
\subsection{Future work}

Although VTONQA provides a comprehensive benchmark for perceptual quality assessment of virtual try-on results, several limitations remain. The current dataset mainly focuses on standard single-garment scenarios, with limited coverage of layered garments, and does not provide dense human annotations (e.g., human parsing maps and pose labels) for fine-grained analysis. Future work will expand the dataset by increasing garment diversity and human subjects, incorporating more complex virtual try-on scenarios, and introducing fine-grained distortion annotations to enable more precise diagnosis of garment deformation, local misalignment, and visual artifacts. In addition, we plan to further standardize image quality and develop a more structured evaluation framework for next-generation virtual try-on models.

\vspace{-15pt}
\section{Conclusion}

In this work, we present VTONQA, the first quality assessment dataset for VTON, comprising 8,132 VTON-generated images and 24,396 MOS annotations across three dimensions: clothing fit, body compatibility, and overall quality. Through subjective assessment of representative VTON methods, we identify key factors affecting try-on quality and reveal limitations of existing objective metrics, emphasizing subjective supervision. We believe VTONQA and its evaluation framework will facilitate perceptually aligned quality assessment methods and reliable VTON algorithms.

\begin{acks}
    This work was supported in part by the National Natural Science Foundation of China under Grants 62225112, 62401365, and in part by the China Postdoctoral Science Foundation under Grant Number BX20250411, 2025M773473.
\end{acks}
\bibliographystyle{ACM-Reference-Format}
\balance
\bibliography{sample-base}

\end{document}